\documentclass{article}

     \usepackage[preprint]{nips_2018}

\usepackage[utf8]{inputenc} 
\usepackage[T1]{fontenc}    
\usepackage{hyperref}       
\usepackage{url}            
\usepackage{booktabs}       
\usepackage{amsfonts}       
\usepackage{nicefrac}       
\usepackage{microtype}      
\usepackage{amsfonts,amsmath,amssymb,array}

\usepackage{makeidx}         
\usepackage{graphicx}        
\usepackage{multicol}        
\usepackage[bottom]{footmisc}

\usepackage{times}
\usepackage{parskip}
\usepackage{subfig}
\usepackage{wrapfig}

\usepackage{amssymb}
\usepackage{epsfig}
\usepackage{amsfonts,amsmath,amssymb,array}
\usepackage{tabularx}   
\usepackage{colortbl}
\usepackage{longtable}
\usepackage{booktabs}
\usepackage{url}
\usepackage{icomma}
\usepackage{pstricks}
\usepackage{longtable}
\usepackage{multirow}
\usepackage{color}
\usepackage{algorithm2e}

\usepackage{subfiles}
\usepackage{float}
\usepackage{mathtools}

\title{Generalized Dilation Neural Networks}

\author{%
  Gavneet Singh Chadha\\
	Automation Technology Group\\
  Department of Electrical Engineering\\
  South-Westphalia University\\
	of Applied Science\\
  Soest, Germany \\
  \texttt{chadha.gavneetsingh@fh-swf.de}\\
  \And
  Jan Niclas Reimann\\
	Automation Technology Group\\
  Department of Electrical Engineering\\
  South-Westphalia University\\
	of Applied Science\\
  Soest, Germany \\
  \texttt{reimann.janniclas@fh-swf.de} \\
  \AND
  Andreas Schwung\\
	Automation Technology Group\\
  Department of Electrical Engineering\\
  South-Westphalia University\\
	of Applied Science\\
  Soest, Germany \\
  \texttt{schwung.andreas@fh-swf.de} \\
}

\begin{document}

\maketitle

\begin{abstract}
Vanilla convolutional neural networks are known to provide superior performance not only in image recognition tasks but also in natural language processing and time series analysis. One of the strengths of convolutional layers is the ability to learn features about spatial relations in the input domain using various parameterized convolutional kernels. However, in time series analysis learning such spatial relations is not necessarily required nor effective. In such cases, kernels which model temporal dependencies or kernels with broader spatial resolutions are recommended for more efficient training as proposed by dilation kernels. However, the dilation has to be fixed a priori which limits the flexibility of the kernels. We propose generalized dilation networks which generalize the initial dilations in two aspects. First we derive an end-to-end learnable architecture for dilation layers where also the dilation rate can be learned. Second we break up the strict dilation structure, in that we develop kernels operating independently in the input space. 
\end{abstract}

\section{Introduction}

Convolutional Neural Networks (CNN)~\citep{Fuku1982,LeCun1989} are known to provide superior performance not only in image recognition tasks but also in speech recognition, natural language processing and time series analysis. One of the strengths of convolutional layers is the ability to learn features about spatial relations in the input domain, which is reasonable in images where the spatial relations are especially fruitful for extraction of representative features. By stacking various convolution layers, the space covered by the convolutions increases from layer to layer such that more and more high level, spatially distributed features are generated. However, for some application, considering wider regions in the image appears to be useful. Beside, considering wider regions might allow for a reduction of the number of layers and active parameters. For input spaces other than images, like time series data, spatial relations are not or just little impactful with regards to the quality of features. Simultaneously, time series features often require considerations of much longer time horizons while subsequent data points do not vary much and hence, provide redundant information which hurt the quality of features. 

\begin{figure}[h]
 \centering
 \includegraphics[width=\columnwidth,keepaspectratio]{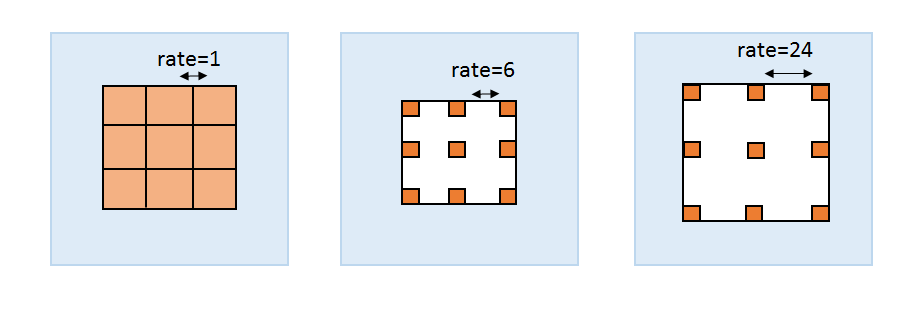}
\caption{Dilation kernels with different dilation rates (where rate=1 is the original convolution operation)~\citep{Chen2017}.}
\label{fig:dilationcoverage}
\end{figure}

To overcome these shortcomings of vanilla CNNs, dilation networks have been proposed in~\citet{Yu2016}. Thereby, dilation refers to the consideration of greater spatial resolutions of the convolution kernels as illustrated in Fig.~\ref{fig:dilationcoverage}. This allows for the coverage of wider spatial regions in the input space and is shown to improve results in various problem domains like image segmentation~\citep{Yu2016}, entity recognition~\citep{Strubell2017}, genomic dependencies~\citep{Gupta2017} and speech recognition~\citep{Sercu2016}. However, the dilation factor which is considered as a hyperparameter has to be fixed a priori.

In this paper, we generalize the previously proposed dilation neural networks in different aspects. First, we relax the original dilation structure which assumes equidistant sampling in the input dimensions. This enables varying sampling in both input dimensions as well as arbitrary patterns for the dilation kernels and hence, allows for a more general selection of relevant input values. Second, we make the dilation kernels learnable, i.e. we not only optimize the weight matrices of each kernels but also the dilation structures making the application of dilation networks more flexible.

\section{Generalized Dilation Neural Networks}

In this section, we introduce generalized dilation layers. We start with the description of dilation networks proposed in~\citet{Yu2016}. Then we generalize the dilation networks in two ways. First, we make the dilation layers end-to-end trainable. Second we relax the dilation structure and propose generalized dilation layers where the dilation structure can take on arbitrary shapes.

\subsection{Dilation Layer}

We give a short introduction into dilation networks previously proposed in~\citet{Yu2016}. We start with the standard convolution operator defined as
\begin{align}
	(F * k)(p) = \sum_{\mathbf{s}+\mathbf{t}=\mathbf{p}} F(\mathbf{s})\cdot k(\mathbf{t}).
\end{align}
In dilated convolution, we have instead
\begin{align}
	(F *_l k)(p) = \sum_{\mathbf{s}+l\mathbf{t}=\mathbf{p}} F(\mathbf{s})\cdot k(\mathbf{t}),
\end{align}
where the term $*_l$ denotes the dilated convolution and $l$ is the dilation factor. Note that the standard convolution operation is obtained for $l=1$. Basically, the above dilated convolution operation results in a modification of the original convolution operator to use its filter parameters in a different way. By changing the dilation factors, the dilated convolution operator applies the same filter (with the same parameters) at different ranges and hence, allows for better multiscale context aggregation. The effect of the dilation operation is illustrated in Fig.~\ref{fig:dilationcoverage} which shows the increasing size of the filter with increasing dilation factor.

\subsection{Generalized dilation layers}

In its original form, the dilation operation, i.e. the dilation factor, is kept fixed during training of the neural network. We allow for more flexibility in that we want to make the dilation factor variable and make this variability trainable.

To this end, we will first derive an alternative representation of the dilation operation as follows. Basically, the dilation operation can be seen as a conventional convolution operation with receptive field size increased from $p\times p$ to $y\times y$ with $p<y$. Correspondingly, an increased weight matrix $W \in \mathbb{R}^{y\times y}$ is applied in which a certain number of weights are fixed to zero a priori such that the active weights (weights unequal zero) sum to $p\times p$. 
Hence, we define vectors $\psi_l \in \{0,1\}^{y}$ and $\psi_r \in \{0,1\}^{y}$ and matrices $\Psi_l$ and $\Psi_r$ where
\begin{align}
	\text{diag}(\Psi_l) = \psi_l, \quad \text{diag}(\Psi_r) = \psi_r.
\end{align}
Than defining a new weight matrix as
\begin{align}
	\tilde{W} = \Psi_l \cdot W \cdot \Psi_r
\end{align}	
and using this weight matrix for convolution provides a generalization to the dilation layer recapitulated in the previous section. In fact, the dilation with active weight size $3\times 3$ and dilation factor two can be obtained by defining $\psi_l=\psi_r=[1\ 0\ 1\ 0\ 1]^T$. Note, that by arbitrarily setting the components of $\psi_l$ and $\psi_r$ to zero and one while at the same time forcing the total number of ones per $\psi_l$ and $\psi_r$ to $p$, respectively, i.e. by imposing the constraints
\begin{align}\label{eq:constraints1}
	\psi^T_l\cdot \mathbf{1} \leq p, \quad \psi^T_r\cdot \mathbf{1} \leq p,
\end{align}
where $\mathbf{1}$ denotes the all-one vector, allows for arbitrary dilation-like patterns.

\begin{figure}[h]
 \centering
 \includegraphics[width=\columnwidth,keepaspectratio]{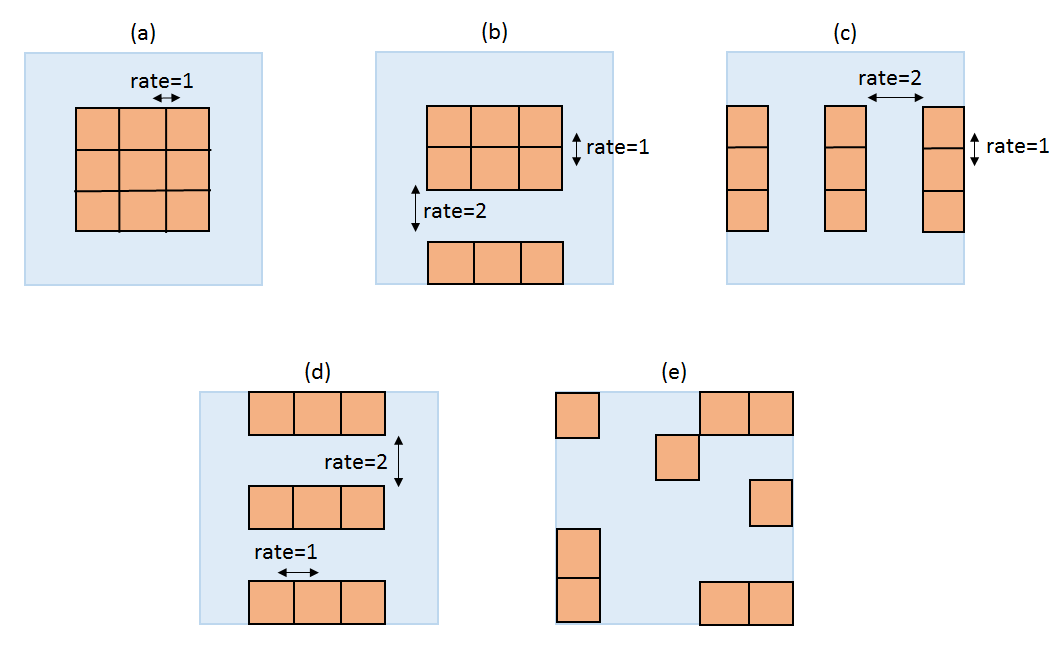}
\caption{Different configurations of the parameters: (a) Original convolution, (b) Dilation with varying dilation rates; (c) Dilation in horizontal dimension only; (c) Dilation in vertical dimension only; (e) Arbitrary dilation kernel.}
\label{fig:dilationpatterns}
\end{figure}

However, we can further generalize the dilation operation by defining a matrix $\Psi \in \{0,1\}^{y\times y}$ and defining the new weight matrix as
\begin{align}
	\tilde{W} = W \odot \Psi
\end{align}	 
where $\odot$ denotes element-wise multiplication. As in the previous case, we again impose constraints on the matrix $\Psi$ as
\begin{align}\label{eq:constraints2}
	\mathbf{1}^T \Psi \cdot \mathbf{1} \leq p^2,
\end{align}
which assures, that the number of weights unequal to zero is less or equal to the kernel size. This definition allows for rather general dilation-like patterns. An illustration of some examples for the different possible patterns is given in Fig.~\ref{fig:dilationpatterns}.
Particularly, some choices of the parameters are useful in practice. The original convolution operation (Fig.~\ref{fig:dilationpatterns} (a)) is obtained by setting $\psi_l=\psi_r=[0\ 1\ 1\ 1\ 0]$ while a dilation with dilation rate 2 yields $\psi_l=\psi_r=[1\ 0\ 1\ 0\ 1]$. However, this equidistant choice of the dilation might not be best to obtain data features in some cases. This is relaxed by setting e.g. $\psi_l=[0\ 1\ 1\ 0\ 1]$ and $\psi_r=[0\ 1\ 1\ 1\ 0]$ resulting in non-equidistant patterns, see Fig.~\ref{fig:dilationpatterns} (b). 

For even more general patterns consider multidimensional time series analysis, where the input space of the CNN consists of the two dimensions time and  sensor input channel. Note that the sensor channels are normally arbitrarily ordered without considering any relations between subsequent channels. Consequently, while a relation in the time dimension is obvious, a relation between subsequent channels is not present in general. In such cases dilation patterns as shown in Fig.~\ref{fig:dilationpatterns} (c) using setting $\psi_l=[0\ 1\ 1\ 1\ 0]$ and $\psi_r=[1\ 0\ 1\ 0\ 1]$ can be beneficial. Similarly, a vertical dilation can be trained (Fig.~\ref{fig:dilationpatterns} (d)). Arbitrary dilation patterns as illustrated in~Fig.~\ref{fig:dilationpatterns} (e) can be obtained by setting 
\begin{align}
	\Psi = \left[\begin{array}{rrrrr}1 & 0 & 0 & 1 & 1 \\0 & 0 & 1 & 0 & 0 \\0 & 0 & 0 & 0 & 1 \\1 & 0 & 0 & 0 & 0 \\1 & 0 & 0 & 1 & 1 \\\end{array}\right]
\end{align}
which is just restricted by the number of data point within the dilation kernel. Higher dilation rates can be obtained by using greater receptive fields.

\subsection{End-to-end training of dilation layers}

So far, we have introduced general dilation layers as an extension of the vanilla dilation layers proposed in~\citet{Yu2016} by defining suitable masking vectors $\psi_l$ and $\psi_r$ and masking matrix $\Psi$, respectively. Hence, as done in convolutional and dilated convolutional neural networks, we can define the masking matrices a priori for each convolutional kernel of the network and keep them fixed during training. However, making these matrices learnable allows for an adjustment of the masking matrices during training and could potentially reduce the number of kernels necessary to the ones essentially needed to represent the given problem. Hence, we define a suitable learning framework for generalized dilation layers in that we make the masking matrices trainable. 
However, optimizing the masking matrix and vectors results in optimizing binary variables which makes the learning problem combinatorical and does not directly allow for end-to-end training. To circumvent this, we follow a similar approach as in~\citet{Trasky2018} in that we define continuous vectors $\tilde{\psi}_l, \tilde{\psi}_r \in \mathbb{R}^{y}$ and matrix $\tilde{\Psi} \in \mathbb{R}^{y\times y}$ which are run through a sigmoidal activation function, i.e. we define
\begin{align}
	\psi_l = \sigma(\tilde{\psi}_l),\quad \psi_r = \sigma(\tilde{\psi}_r),\quad \Psi = \sigma(\tilde{\Psi}).
\end{align}
Using this reparameterization, the masking vector and matrix are bounded to the interval $[0, \ 1]$ and due to the characteristics of the sigmoid function will tend to its boundaries as training proceeds.

Using the above reparameterization we end up with a trainable structure with the standard set of network parameters $\omega$ consisting of the kernel weights of the convolution/dilation kernels and the parameters of the fully connected layers and the parameter vectors or matrices $\Psi$. This can be trained by backpropagation using stochastic gradient descent on the standard loss functions $L_s(\omega,\Psi)$ used for CNNs like cross entropy loss or mean squared error. However, we have to consider the additional constraints on the parameters $\Psi$ as given in Eq.~\eqref{eq:constraints1} and~\eqref{eq:constraints2}. These constraint have to be fulfilled during training and hence, impose hard constraints.

In general, various approaches exist to incorporate hard constraints in stochastic gradient descent algorithms, namely barrier functions, projection methods and active set methods. In this work we propose to use differentiable barrier functions which contribute to the loss function only if the constraints are not satisfied. To this end, we describe the constraints by inequality constraints of the form $f_c(\Psi)\leq 0$ such that the barrier loss function $L_c{\Psi}$ yields 
\begin{align}
	L_c(\Psi) = \sum_{f^i_c\in \mathcal{C}} b_c(f^i_c(\Psi)),
\end{align}
where $\mathcal{C}$ is the set of inequality constraints and $b_c$ is an arbitrary barrier function chosen to be an exponential barrier function
\begin{align}
	b_c(x) =  e^{10\cdot \left(x-0.5\right)}+\alpha \cdot x,
\end{align}
where we experiment with different slope parameters $\alpha \in [-0.1,\ 0.1]$. Finally, the loss function for end-to-end training of generalized dilation networks yields
\begin{align}
	L(\omega,\Psi) = L_s(\omega,\Psi) + \mu \cdot L_c(\Psi).
\end{align}
where $\mu > 0$ is a parameter to account for the continuous approximation of the barrier function.

\section{Related Work}

Convolution with dilated filters have first been introduced in~\citet{Hol1989} and~\citet{Shensa1992} in the context of wavelet decomposition. Dilated convolutions have been initially presented in~\citet{Yu2016} to allow for multi-scale context aggregation without downsampling of the resolution. Since then, dilation networks have been used in semantic segmentation methods due to their ability to capture large context while preserving fine details. In~\citet{Chen2016a}, large dilation factor are used in the Deeplab model~\citep{Chen2014} to provide large context, which results in improved performance. This is further enhanced in~\citet{Chen2016} by using atrous spatial pyramid pooling (ASPP), i.e. multi-level dilated convolutions, improving the results by leveraging local and wide context information. In~\citet{Sercu2016}, time dilated convolutions are used for dense prediction on sequences for speech recognition while modeling long-distance genomic dependencies with dilated convolutions are reported in~\citet{Gupta2017}. Iterated dilated convolutions are applied in~\citet{Strubell2017} for entity recognition. \citet{Strubell2017} use dilated convolutions for improving the performance of variational autoencoders for text modeling. Dilated residual networks are introduced in~\citet{Yu2017}, considerably improving vanilla residual networks~\citep{He2016} in image classification and segmentation. However, in these works the dilation parameters are fixed and hand-tuned a priori. In contrast, we keep the dilation operation end-to-end trainable for each channel. 

Such training of dilation factors have been presented in~\citet{He2017} where a dilation factor is trained for each channel of the convolution layer. Furthermore, the dilation factor is defined in $\mathbb{R}$ instead of $\mathbb{Z}_{+}$ as in the original derivation. This comes at the cost of calculating the output map by means of a bilinear transformation in the regular case of fractional dilation factors. Furthermore, the structure of the dilation is fixed. More general structures are allowed by active convolutions~\citet{Jeon2017}, however with low deviations from the original convolution kernel. Deformable convolutional networks as introduced in~\citet{Dai2017} and further improved in~\citet{Zhu2018} allowing for similar convolution structures as proposed by our approach. There, each point in the convolution grid is augmented with a learnable real valued offset. As in~\citet{He2017}, a bilinear transformation is needed before applying the convolution. Contrary, we hold on to integer dilation factors while allowing for arbitrary structures of the kernel using end-to-end trainable sparse masking matrices. Due to that, we end up in the exact grid position without the need for a bilinear transformation which can be problematic in time series analysis, especially with discrete or binary time series. Furthermore, the context of this paper and the application of the network architecture is mainly focused on multidimensional time series data analysis and not on segmentation problems. Application of dilation to time series analysis is provided in WaveNets~\citep{vdO2016} but with fixed dilation factors only.

\section{Experiments}

We leave results and experiments on image classification, image segmentation and time series clustering and classification to future work.

\section{Conclusion}

We presented generalized dilation neural networks, a NN architecture based on CNNs augmented with dilated filters. We provided extensions to this framework in two ways. First we make the fixed dilation filters learnable by introducing an alternative representation of the dilation operation using masking vectors or matrices which can be made end-to-end trainable. Second, we generalize the fixed structure of dilation kernels to arbitrary structures, allowing for an arbitrary coverage of the input space with more effective image section selection (in particular for initial layers) and dilation networks.

\end{document}